\documentclass{article}

\usepackage[preprint]{neurips_2026}
\usepackage{tabularx}

\usepackage[utf8]{inputenc}
\usepackage[T1]{fontenc}
\usepackage{graphicx}
\usepackage{multirow}
\usepackage{amsmath,amssymb,amsfonts}
\usepackage{amsthm}
\usepackage{mathrsfs}
\usepackage[title]{appendix}
\usepackage{xcolor}
\usepackage{textcomp}
\usepackage{booktabs}
\usepackage{algorithm}
\usepackage{algorithmic}
\usepackage{url}
\usepackage{hyperref}

\theoremstyle{plain}

\theoremstyle{definition}

\usepackage[utf8]{inputenc} 
\usepackage[T1]{fontenc}    
\usepackage{hyperref}       
\usepackage{url}            
\usepackage{booktabs}       
\usepackage{amsfonts}       
\usepackage{nicefrac}       
\usepackage{microtype}      
\usepackage{xcolor}         

\title{ProWAFT: A ROMA-LPD Instance for Workload-Aware and Dynamic Fault Tolerance in FPGA-Based CNN Accelerators}

\author{%
Xinxin Chen \\
Department of Computer Science\\
University of Chinese Academy of Sciences\\
\And
Haoran Qiao \\
Department of Computer Science\\
University of Chinese Academy of Sciences\\
\And
Yiming Guo \\
Department of Computer Science\\
University of Chinese Academy of Sciences\\
\And
Kecheng Luo \\
Department of Computer Science\\
University of Chinese Academy of Sciences\\
\And
Siyuan Feng \\
Department of Computer Science\\
University of Chinese Academy of Sciences\\
\And
Jingwen Ma \\
Department of Computer Science\\
University of Chinese Academy of Sciences\\
}

\begin{document}

\maketitle

\begin{abstract}

SRAM-based FPGAs provide an attractive platform for energy- and latency-constrained CNN inference at the network edge, yet transient faults can lead to silent errors that compromise reliability. Always-on redundancy (e.g., full TMR) improves correctness but incurs substantial performance and energy overhead, while reactive recovery may introduce unacceptable latency on the critical path. We propose \textbf{ProWAFT}, a proactive workload-aware fault-tolerance framework for FPGA-based CNN accelerators that uses partial reconfiguration to selectively apply TMR across reconfigurable partitions. ProWAFT quantifies workload criticality, models fault propagation and reconfiguration overhead, and selects configurations that minimize a composite objective over latency, energy, and reliability risk. Implemented on a Xilinx Zynq UltraScale+ ZCU104 platform with six reconfigurable regions and evaluated on a 500-task trace derived from ResNet-18, MobileNetV2, and EfficientNet-Lite under time-varying SEU injection, ProWAFT achieves lower composite cost than static TMR and reactive reconfiguration while maintaining high task success rate and near-baseline throughput with low online decision overhead.

\end{abstract}

\section{Introduction}
\label{sec:introduction}

Convolutional neural networks (CNNs) are now routinely deployed on edge platforms for perception and monitoring tasks where latency and energy are tightly constrained. FPGA-based accelerators are a practical option in this regime: they offer domain-specific parallelism while retaining post-deployment flexibility through reconfiguration. As deployments scale and move into harsher operating environments, however, reliability becomes a limiting factor rather than a secondary concern.\cite{safework2025r1}

\begin{figure}
    \centering
    \includegraphics[width=1\linewidth]{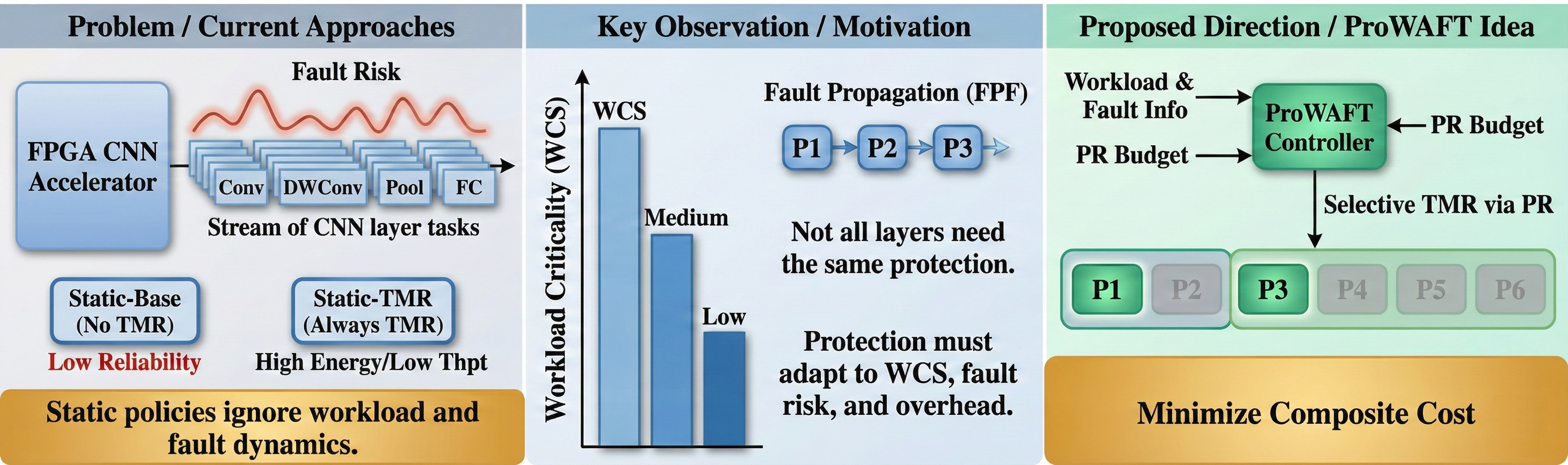}
    \caption{Motivation for ProWAFT. Static approaches fail by ignoring dynamic fault risks and varied workload criticality. ProWAFT addresses this by proactively adapting protection based on Workload Criticality (WCS), real-time fault risk, and reconfiguration overhead. The controller manages selective TMR via Partial Reconfiguration (PR) to balance performance, energy, and reliability, minimizing composite cost.}
    \label{fig:motivation}
\end{figure}

SRAM-based FPGAs are vulnerable to transient disturbances that can corrupt configuration bits or on-chip computation, leading to silent errors in CNN inference. The difficulty is that fault impact is highly non-uniform. Both the fault risk (e.g., varying with time and operating conditions) and the workload sensitivity (e.g., layer type, precision, and error-propagation behavior) can change substantially across an execution trace. Figure~\ref{fig:motivation} summarizes the resulting mismatch: a fixed protection level assumes a stationary environment and a homogeneous workload, neither of which holds in practice.\cite{wen2026devil}

A common mitigation is to apply triple modular redundancy (TMR) statically, which improves robustness but also increases resource usage and power, and can reduce throughput on constrained devices. At the other extreme, running without redundancy preserves efficiency but exposes the system to unacceptable error rates when fault risk increases. Reactive recovery schemes that reconfigure only after detecting a fault reduce steady-state overhead, but they pay the cost on the critical path and can violate latency constraints. These trade-offs indicate that reliability should be treated as a runtime decision rather than a one-time design choice.\cite{gao2025exploring}

This work presents \textbf{ProWAFT}, a proactive workload-aware fault-tolerance framework for FPGA-based CNN accelerators. ProWAFT uses runtime telemetry (workload features and estimated per-partition fault probabilities) together with offline-characterized partial reconfiguration (PR) overhead to decide when and where to enable protection. The approach quantifies workload criticality, models fault propagation and recovery risk across reconfigurable partitions, and selects a configuration that minimizes a composite objective spanning performance, energy, and reliability while accounting for PR cost. In our implementation on a Zynq UltraScale+ ZCU104 platform with six reconfigurable regions and a 500-task CNN workload trace, ProWAFT reduces composite cost relative to static and reactive baselines while keeping decision overhead small.\cite{li-etal-2025-treble}

\paragraph{Contributions.}
\begin{itemize}
    \item We formulate proactive, workload-aware fault tolerance for FPGA-based CNN accelerators as a runtime configuration problem that enables \emph{selective TMR} via partial reconfiguration, explicitly trading off latency, energy, and reliability under reconfiguration constraints.
    \item We introduce a lightweight modeling stack for decision-making, including a \emph{Workload Criticality Score (WCS)} and a fault-propagation-based risk formulation (FPF/RRS), and integrate these components into a composite cost used for configuration selection.
    \item We implement ProWAFT on a ZCU104 platform and evaluate it under a time-varying SEU injection model using diverse CNN-layer workloads. Results show improved composite cost and adaptive protection behavior, with low online decision overhead and quantified PR costs.
\end{itemize}

\section{Related Work}
\label{sec:related}

\subsection{FPGA Soft Errors}
SRAM-based FPGAs are susceptible to transient faults such as single-event upsets (SEUs), which may corrupt configuration bits and lead to silent malfunctions. Prior work has explored mitigation across the stack, including configuration scrubbing (periodic or adaptive), protection of state with ECC/parity, and hardware redundancy such as TMR/DWC . These techniques can substantially improve reliability, but their overhead is non-trivial and often workload- and platform-dependent. In edge inference, where power and latency margins are limited, always-on redundancy is frequently too expensive, while purely reactive recovery can incur unacceptable latency spikes.\cite{tang2022few}

\subsection{CNN Reliability}
A growing body of work studies the reliability of CNN/DNN inference under faults via fault injection, statistical error models, and vulnerability analysis. A consistent observation is that fault impact is not uniform: different layer types, precisions, and activation behaviors exhibit different sensitivities, and errors may amplify as they propagate through the network. This has motivated selective protection and lightweight detection strategies for DNN accelerators. However, many existing approaches either assume a fixed protection plan chosen offline or do not explicitly model how faults propagate through dependent accelerator stages in a partitioned design.\cite{liu2023spts}

\subsection{Partial Reconfiguration}
Partial reconfiguration (PR) has been widely used to time-multiplex accelerator functions, specialize datapaths to changing models, and adapt resource allocation at runtime. PR also enables adaptive reliability by switching between baseline and hardened variants (e.g., TMR-protected modules) when conditions warrant  . Existing PR-based fault-tolerance schemes are often reactive (triggered after an error) or rely on fixed thresholds, and they may not explicitly account for the time/energy overhead of PR in the decision objective. This can lead to protection being applied too late, too broadly, or too frequently.
\cite{tang2022optimal}

\subsection{Runtime Policies}
Runtime management under competing objectives (latency, energy, and reliability) has been addressed using rule-based switching, heuristic controllers, and optimization/learning-based policies such as MDP or RL formulations. These methods demonstrate the value of formal decision-making, but they typically adopt coarse reliability surrogates and do not incorporate workload-dependent fault criticality together with fault propagation and PR overhead in a unified cost model. ProWAFT differs by combining workload criticality scoring with propagation-aware risk modeling and an explicit PR cost term, enabling proactive selective TMR decisions that track both workload variation and time-varying fault risk.
\cite{feng2024docpedia}

\begin{figure}
    \centering
    \includegraphics[width=1\linewidth]{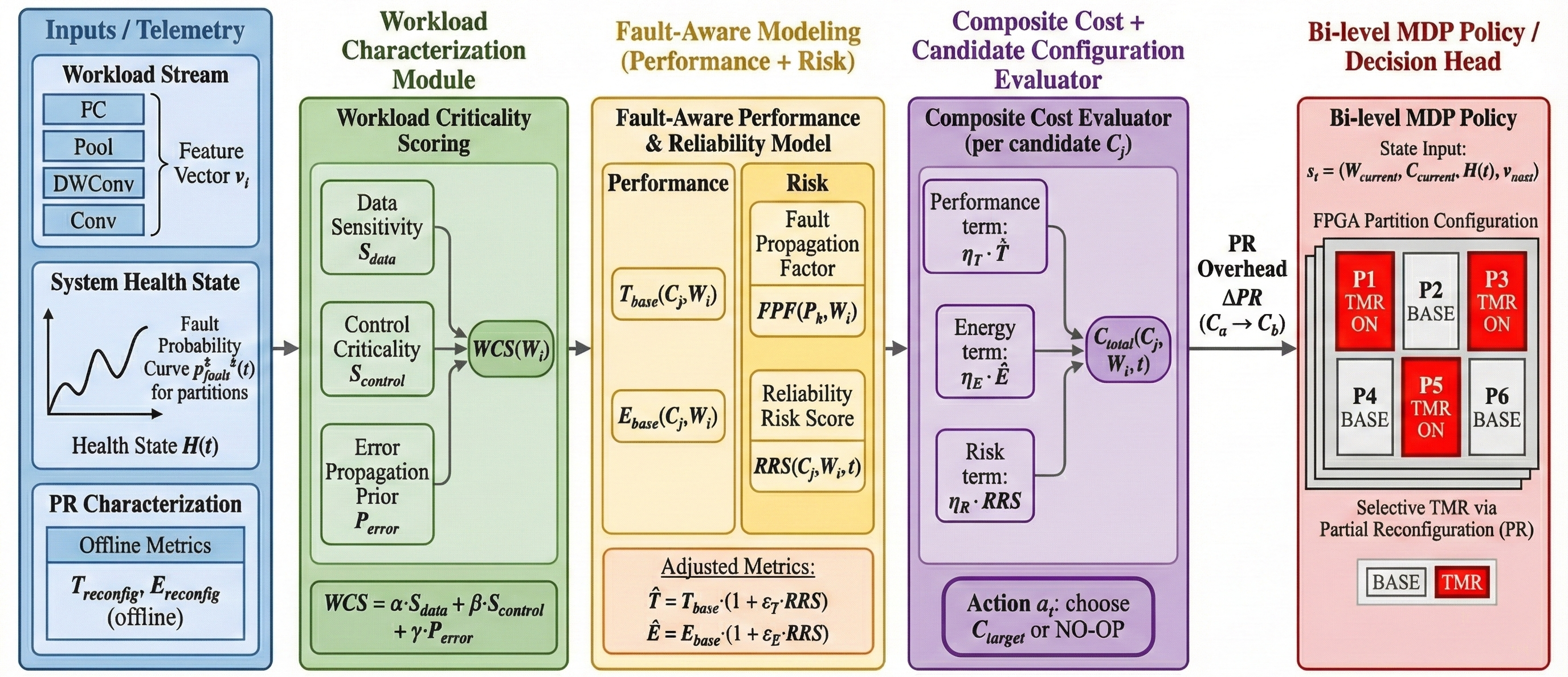}
    \caption{Architectural overview of the ProWAFT pipeline. Input telemetry from workloads and system health is processed sequentially through four stages: (1) Workload Criticality Scoring (WCS); (2) Fault-Aware Performance \& Risk Modeling; (3) Composite Cost Evaluation based on weighted metrics and PR overhead; and finally, (4) a Bi-level MDP Policy that determines the optimal proactive selective TMR action for the FPGA partitions.}
    \label{fig:overview}
\end{figure}

\section{Methodology: Proactive Workload-Aware Fault Tolerance (ProWAFT) for CNN Accelerators}
\label{sec:method}

This section presents \textbf{ProWAFT}, a runtime framework that enables workload-adaptive fault tolerance for FPGA-based CNN accelerators via Partial Reconfiguration (PR). ProWAFT decides \emph{when} and \emph{where} to enable protection (e.g., TMR variants) by jointly considering (i) workload characteristics, (ii) estimated partition-level fault risk, and (iii) PR overhead (Fig.~\ref{fig:overview}).

\subsection{System Model and Problem Formulation}
We model the accelerator as a set of $K$ reconfigurable partitions $\mathcal{P}=\{P_1,\dots,P_K\}$. Each partition $P_k$ can be configured with one of $N_k$ pre-synthesized hardware functions from a library $\mathcal{F}_k$. In our setting, $\mathcal{F}_k$ includes both baseline and protected (e.g., TMR) variants for supported kernels. A \textbf{system configuration} is denoted by $C_j\in\mathcal{C}$.\cite{zhao2024multi}
\cite{zhao2024harmonizing}

The accelerator processes a stream of workloads $\mathcal{W}=\{W_1,W_2,\dots\}$, where each $W_i$ corresponds to a CNN layer (or a small group of layers) described by a feature vector $\mathbf{v}_i$ (e.g., operator type, input shape, precision, batch size). We denote by $\mathcal{C}(W_i)\subseteq\mathcal{C}$ the set of \textbf{feasible configurations} for workload $W_i$, i.e., configurations that implement the required kernels and precision for $W_i$.
\cite{wang2025pargo}
\cite{tang2023character}

The system health state at time $t$ is represented by $\mathbf{H}(t)=\{p^{fault}_1(t),\dots,p^{fault}_K(t)\}$, where $p^{fault}_k(t)$ is the estimated transient fault probability for partition $P_k$. PR is performed between workloads and incurs time/energy overhead; we model a constrained reconfiguration budget $B_{PR}$ (time and/or energy) over an execution window.
\cite{sun2025attentive}
\cite{lu2024bounding}

\noindent\textbf{Objective.} For each incoming workload $W_i$, given the current configuration $C_{current}$, health state $\mathbf{H}(t)$, and remaining budget $B_{PR}$, ProWAFT selects an action (reconfigure to some $C_j\in\mathcal{C}(W_i)$ or stay) to minimize an expected composite cost that captures latency, energy, and reliability risk while accounting for PR overhead.
\cite{zhao2025tabpedia}
\cite{tang2024mtvqa}
\cite{tang2024textsquare}

\subsection{Workload Characterization and Criticality Scoring}
We define a \textbf{Workload Criticality Score (WCS)} to quantify how sensitive $W_i$ is to hardware faults:
\begin{equation}
    WCS(W_i)=\alpha\cdot \mathcal{S}_{data}(W_i)+\beta\cdot \mathcal{S}_{control}(W_i)+\gamma\cdot \mathcal{P}_{error}(W_i),
    \label{eq:wcs}
\end{equation}
where $\alpha+\beta+\gamma=1$ and each term is normalized to $[0,1]$.
\cite{shan2024mctbench}
\cite{feng2023unidoc}
\cite{tang2022youcan}

\begin{itemize}
    \item $\mathcal{S}_{data}(W_i)$ (\textbf{data sensitivity}) is derived from the entropy of the output activation distribution. In practice, we estimate it using offline profiling on a calibration set and store layer-wise statistics as a lookup table indexed by operator type/shape/precision.
    \item $\mathcal{S}_{control}(W_i)$ (\textbf{control criticality}) flags workloads on conditional paths (0/1). For standard feed-forward CNNs without dynamic control flow, this term is set to 0.
    \item $\mathcal{P}_{error}(W_i)$ (\textbf{error propagation likelihood}) is obtained from lightweight pre-characterization (e.g., single-bit fault injection) and stored as a compact table or regression model. It approximates the probability that a fault in the current workload causes a task-level critical error at the output.
\end{itemize}

\subsection{Fault-Aware Performance and Reliability Modeling}
For a candidate configuration $C_j\in\mathcal{C}(W_i)$, ProWAFT predicts fault-free performance/energy, and then estimates reliability risk under $\mathbf{H}(t)$.

\textbf{1) Baseline (fault-free) metrics.}
\begin{align}
    T_{base}(C_j,W_i) &= \sum_{k=1}^{K}\frac{Ops_k(W_i)}{f_k\cdot PE_k(C_j)} + T_{comm}(C_j,W_i), \label{eq:t_base}\\
    E_{base}(C_j,W_i) &= \sum_{k=1}^{K} P^{dyn}_k(C_j)\cdot \frac{Ops_k(W_i)}{f_k\cdot PE_k(C_j)} + P_{static}(C_j)\cdot T_{base}(C_j,W_i).
    \label{eq:e_base}
\end{align}
Here $Ops_k(W_i)$ denotes the operations mapped to $P_k$ for $W_i$, $f_k$ is the clock frequency, $PE_k(C_j)$ is the number of active processing elements, $P^{dyn}_k(C_j)$ is dynamic power, and $P_{static}(C_j)$ is static power of the active configuration.
\cite{fu2024ocrbench}
\cite{guo2025seed1}
\cite{wang2025vision}

\textbf{2) Fault propagation and risk.}
We define a \textbf{Fault Propagation Factor (FPF)} for each partition:
\begin{equation}
    FPF(P_k,W_i,C_j)=WCS(W_i)\cdot \lambda_k(C_j)\cdot Fanout(P_k,C_j),
    \label{eq:fpf}
\end{equation}
where $\lambda_k(C_j)\in[0,1]$ is the utilization ratio of $P_k$ under $C_j$, and $Fanout(P_k,C_j)$ is the number of downstream partitions that consume $P_k$'s outputs in the datapath induced by $C_j$.
\cite{wang2025wilddoc}
\cite{feng2025dolphin}
\cite{lu2025prolonged}
\cite{fei2025advancing}
We then compute a \textbf{dimensionless Reliability Risk Score}:
\begin{equation}
    RRS(C_j,W_i,t)=\frac{1}{Z}\sum_{k=1}^{K} p^{fault}_k(t)\cdot FPF(P_k,W_i,C_j)\cdot \rho_k,
    \label{eq:rrs}
\end{equation}
where $\rho_k\in[0,1]$ is a normalized severity weight reflecting how costly a fault in $P_k$ is (e.g., based on recovery difficulty or observed output impact), and $Z$ is a normalization constant chosen so that $RRS\in[0,1]$ over the considered candidate set.

To couple risk with efficiency metrics, we use fault-aware predictions:
\[
\hat{T}=T_{base}\cdot(1+\epsilon_T\cdot RRS),\qquad
\hat{E}=E_{base}\cdot(1+\epsilon_E\cdot RRS),
\]
where $\epsilon_T$ and $\epsilon_E$ are obtained from offline fault characterization.

\textbf{3) PR overhead.}
Switching from $C_a$ to $C_b$ incurs:
\begin{equation}
    \Delta_{PR}(C_a\rightarrow C_b)=\omega_T\cdot T_{reconfig}(C_a,C_b)+\omega_E\cdot E_{reconfig}(C_a,C_b),
    \label{eq:pr_overhead}
\end{equation}
with $T_{reconfig}$ and $E_{reconfig}$ characterized offline.

\subsection{Policy and Online Decision Rule}
ProWAFT can be viewed as a sequential decision problem: at each step, the controller selects a reconfiguration action based on $(W_i,C_{current},\mathbf{H}(t))$ and the remaining budget. We define the per-workload composite cost for a candidate configuration $C_j$ as:
\begin{equation}
    \mathcal{C}_{total}(C_j,W_i,t)=\eta_T\cdot \tilde{T}(C_j,W_i,t)+\eta_E\cdot \tilde{E}(C_j,W_i,t)+\eta_R\cdot RRS(C_j,W_i,t),
    \label{eq:composite_cost}
\end{equation}
where $\eta_T,\eta_E,\eta_R$ are application weights and $\tilde{T},\tilde{E}$ are normalized latency/energy terms (e.g., normalized to the fault-free static-base reference for the same workload) to keep the objective dimensionless and comparable across workloads.

In our implementation, we use a low-overhead receding-horizon controller: for each incoming $W_i$, we evaluate a finite candidate set $\mathcal{C}(W_i)$ (or a pruned subset) and choose the configuration that minimizes the immediate objective plus PR overhead, subject to the remaining budget:
\begin{equation}
    J(C_j)=\mathcal{C}_{total}(C_j,W_i,t)+\mathbb{I}[C_j\neq C_{current}]\cdot \Delta_{PR}(C_{current}\rightarrow C_j).
    \label{eq:reward}
\end{equation}
The selected action is $a_t=\arg\min_{C_j\in\mathcal{C}(W_i)} J(C_j)$, with infeasible actions filtered out when $\Delta_{PR}$ violates the current budget.

\section{Experiments and Evaluation}
\label{sec:experiments}

This section evaluates \textbf{ProWAFT} on a real FPGA platform under a controlled, time-varying fault injection setting. We answer four research questions:
(1) \textbf{Effectiveness}: Does ProWAFT improve the overall performance--energy--reliability trade-off compared to static and reactive baselines?
(2) \textbf{Adaptivity}: Does ProWAFT adapt protection decisions to workload criticality and fault risk?
(3) \textbf{Overhead}: What is the runtime overhead of decision-making and PR, and how does it compare to reactive recovery?
(4) \textbf{Component contribution}: How much does each ProWAFT component contribute?

Unless otherwise stated, all methods are evaluated on the same 500-task trace and the same fault injection schedule (fixed random seed) for fair comparison.

\subsection{Experimental Setup}
\label{subsec:exp_setup}

\subsubsection{Platform}
We implement ProWAFT on a Xilinx Zynq UltraScale+ ZCU104 platform. The FPGA fabric is partitioned into $K{=}6$ reconfigurable regions. Each region can host one of the pre-synthesized accelerator variants (baseline and TMR-protected). Table~\ref{tab:platform_specs} summarizes the platform and accelerator library.

\begin{table}[htbp]
\centering
\caption{Experimental Platform Specifications and Accelerator Library}
\label{tab:platform_specs}
\begin{tabular}{@{}ll@{}}
\toprule
\textbf{Component} & \textbf{Specification} \\
\midrule
FPGA Platform & Xilinx Zynq UltraScale+ ZCU104 \\
Processor & ARM Cortex-A53 Quad-core @ 1.5GHz \\
FPGA Fabric & Kintex UltraScale+ (504K Logic Cells) \\
Reconfigurable Partitions ($K$) & 6 \\
\addlinespace
\multicolumn{2}{@{}l@{}}{\textbf{Accelerator Library ($\mathcal{F}_k$)}} \\
\addlinespace[0.2em]
8-bit INT Convolution Engine (CE) & Baseline convolution accelerator \\
Max-Pooling Unit (PU) & Pooling operation accelerator \\
BatchNorm-Activation Unit (BAU) & BatchNorm and activation accelerator \\
CE-TMR / PU-TMR / BAU-TMR & TMR-protected versions (triplicated logic) \\
\bottomrule
\end{tabular}
\end{table}

\subsubsection{Workload trace}
We construct a workload trace $\mathcal{W}$ with 500 tasks derived from representative CNNs (ResNet-18, MobileNetV2, and EfficientNet-Lite). Each task corresponds to a supported layer type (Conv2D, DepthwiseConv2D, Pooling, FC) with varying input dimensions. Table~\ref{tab:workload_stats} summarizes the trace.

\begin{table}[htbp]
\centering
\caption{Workload Trace Characteristics}
\label{tab:workload_stats}
\begin{tabular}{@{}lc@{}}
\toprule
\textbf{Parameter} & \textbf{Value/Range} \\
\midrule
Total Workloads & 500 \\
CNN Models & ResNet-18, MobileNetV2, EfficientNet-Lite \\
Layer Types & Conv2D, DepthwiseConv2D, Pooling, FC \\
Input Dimensions & $32{\times}32$ to $224{\times}224$ \\
Workload Criticality ($WCS$) Range & 0.15 -- 0.85 \\
\bottomrule
\end{tabular}
\end{table}

\subsubsection{Fault injection and detection}
We inject transient faults as single-event upsets (SEUs) targeting configuration memory bits. The per-partition fault probability $p_k^{fault}(t)$ varies over time (sinusoidal component plus random baseline) within $[0.001, 0.01]$. Table~\ref{tab:fault_model} lists the fault model settings. Faults are detected using a lightweight parity check; upon detection, the reactive baseline triggers recovery by switching to protected configurations.

\begin{table}[htbp]
\centering
\caption{Fault Injection Model Parameters}
\label{tab:fault_model}
\begin{tabular}{@{}lc@{}}
\toprule
\textbf{Parameter} & \textbf{Value} \\
\midrule
Fault Type & Single-Event Upset (SEU) \\
Injection Target & Configuration memory bits \\
Fault Probability $p_k^{fault}(t)$ Range & 0.001 -- 0.01 \\
Probability Variation & Sinusoidal + random baseline \\
Fault Detection Method & Lightweight parity check \\
\bottomrule
\end{tabular}
\end{table}

\subsubsection{Metrics and baselines}
We compare ProWAFT against three baselines: \textbf{Static-Base} (always baseline accelerators), \textbf{Static-TMR} (always TMR variants), and \textbf{Reactive-Reconfig (RR)} (reconfigure to TMR only after fault detection). We report composite cost $C_{total}$ (Eq.~\ref{eq:composite_cost}) as the primary metric, together with normalized throughput, system energy, task success rate, and PR overhead. Table~\ref{tab:metrics_baselines} summarizes metrics and baselines. Unless otherwise specified, we use $(\eta_T,\eta_E,\eta_R)=(0.4,0.3,0.3)$.

\begin{table}[htbp]
\centering
\caption{Evaluation Metrics and Baseline Methods}
\label{tab:metrics_baselines}
\begin{tabularx}{\linewidth}{@{}p{0.22\linewidth}p{0.28\linewidth}X@{}}
\toprule
\textbf{Category} & \textbf{Item} & \textbf{Description} \\
\midrule
\textbf{Metrics} & Composite Cost ($C_{total}$) & Primary metric from Eq.~\ref{eq:composite_cost} \\
\textbf{Metrics} & Normalized Throughput & Workloads/sec, normalized to Static-Base \\
\textbf{Metrics} & System Energy (J) & Total energy consumed during the trace \\
\textbf{Metrics} & Task Success Rate (\%) & Workloads completed without critical error \\
\textbf{Metrics} & Reconfiguration Overhead & Time/energy spent on PR events \\
\addlinespace
\textbf{Baselines} & Static-TMR & All partitions use TMR-protected accelerators \\
\textbf{Baselines} & Static-Base & All partitions use baseline (non-TMR) accelerators \\
\textbf{Baselines} & Reactive-Reconfig (RR) & Reconfigure to TMR only after fault detection \\
\addlinespace
\textbf{Weights} & $(\eta_T,\eta_E,\eta_R)$ & 0.4, 0.3, 0.3 (balanced profile) \\
\bottomrule
\end{tabularx}
\end{table}

\subsubsection{Measurement methodology}
We measure end-to-end execution time over the full trace on the ARM processor (PS) and compute normalized throughput as workloads/sec normalized to Static-Base. PR time is measured by timestamping the start/end of each PR event. System energy is obtained by integrating measured power over the trace duration using the same procedure across all methods. For success rate, a workload is counted as successful if it completes and its output passes the defined correctness check (golden-reference comparison or application-defined tolerance).

\subsection{Results}
\label{subsec:exp_results}

\subsubsection{Trace profile visualization}
\begin{figure}[t]
\centering
\includegraphics[width=\linewidth]{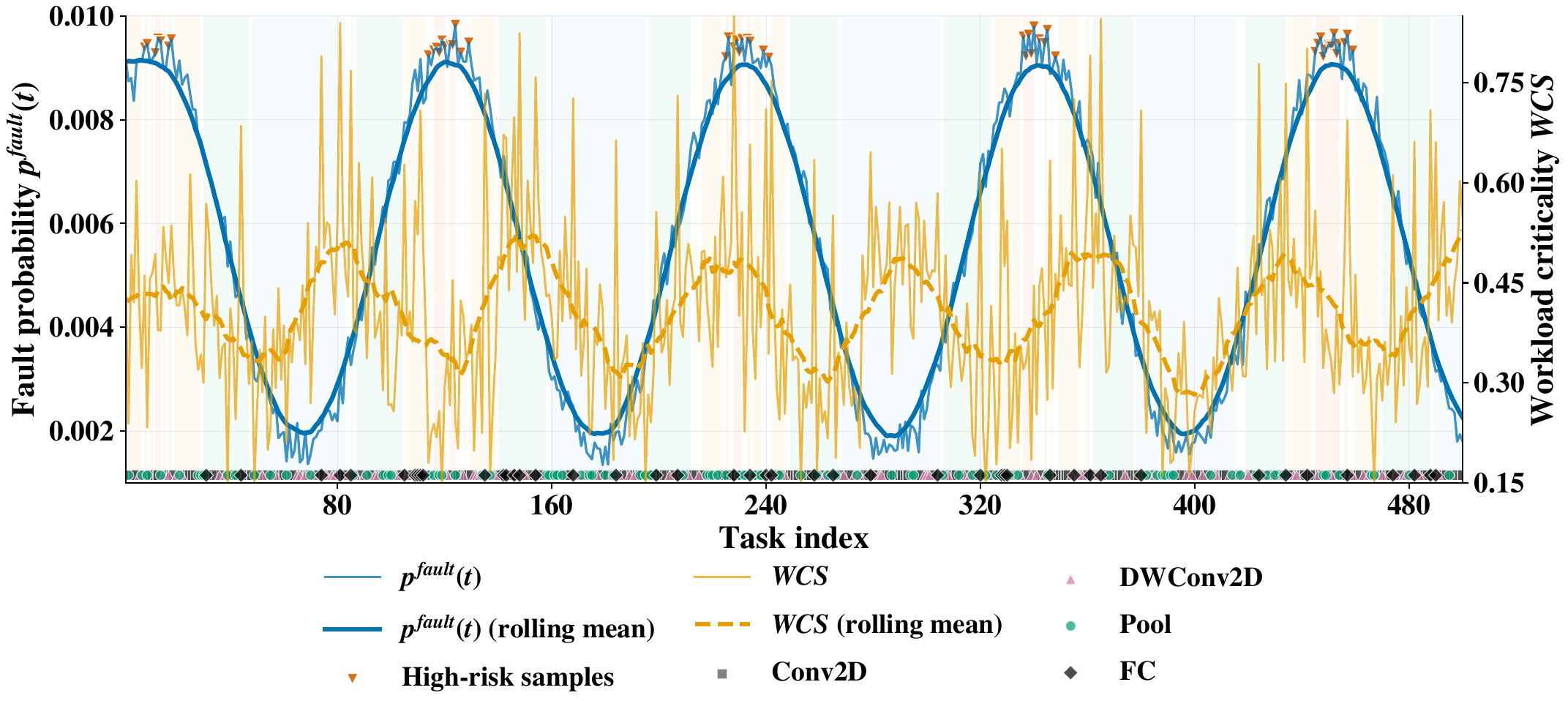}
\caption{Workload trace and time-varying fault risk used in evaluation. Top: representative (or average) partition fault probability $p_k^{fault}(t)$ over the 500-task trace. Bottom: per-task workload criticality score (WCS).}
\label{fig:trace_profile}
\end{figure}

To illustrate the evaluation setting, Fig.~\ref{fig:trace_profile} visualizes the time-varying fault risk and workload criticality over the 500-task trace (used in all experiments).

\subsubsection{Overall effectiveness (RQ1)}
Table~\ref{tab:overall_results} reports overall performance. Static-Base achieves the highest throughput but suffers the lowest success rate. Static-TMR achieves perfect success but pays a large throughput and energy penalty. RR improves success rate but still incurs substantial overhead.

ProWAFT achieves the best trade-off: compared to Static-TMR, it improves normalized throughput from 0.61 to 0.89 (+45.9\%), reduces energy from 302.5\,J to 210.7\,J (-30.3\%), and maintains a high success rate (98.8\%, 1.2 percentage points lower than Static-TMR). Compared to RR, ProWAFT improves throughput by 18.7\%, reduces energy by 17.1\%, improves success rate by 2.4 percentage points, and reduces $C_{total}$ from 0.78 to 0.54 (30.8\% relative reduction).

\begin{table}[htbp]
\centering
\caption{Overall Performance Comparison Across All Methods}
\label{tab:overall_results}
\begin{tabular}{@{}lcccc@{}}
\toprule
\textbf{Method} & \textbf{Norm. Throughput} & \textbf{Energy (J)} & \textbf{Success Rate (\%)} & $\mathbf{C_{total}}$ \\
\midrule
Static-Base & 1.00 & 185.2 & 78.0 & 0.72 \\
Static-TMR & 0.61 & 302.5 & 100.0 & 0.81 \\
Reactive-Reconfig & 0.75 & 254.1 & 96.4 & 0.78 \\
\textbf{ProWAFT (Ours)} & \textbf{0.89} & \textbf{210.7} & \textbf{98.8} & \textbf{0.54} \\
\bottomrule
\end{tabular}
\end{table}

\begin{figure}[t]
\centering
\includegraphics[width=0.92\linewidth]{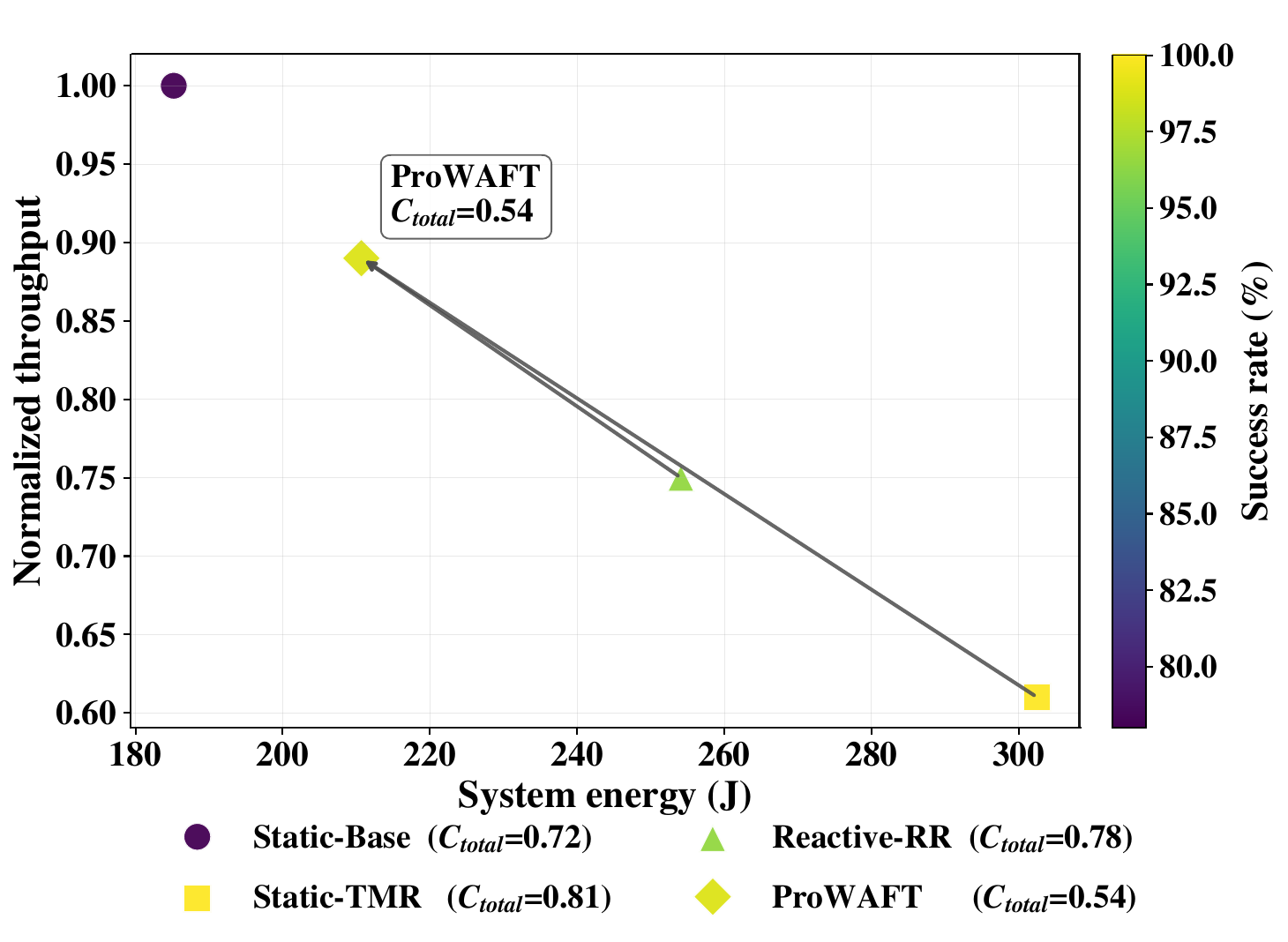}
\caption{Energy--throughput trade-off with reliability annotation (e.g., point label or marker size indicates success rate). ProWAFT achieves a favorable operating point compared to static and reactive baselines.}
\label{fig:tradeoff}
\end{figure}

We further visualize the operating points in Fig.~\ref{fig:tradeoff}, where ProWAFT lies in a favorable region compared to static and reactive baselines.

\subsubsection{Adaptivity (RQ2)}
\begin{figure}[t]
\centering
\includegraphics[width=\linewidth]{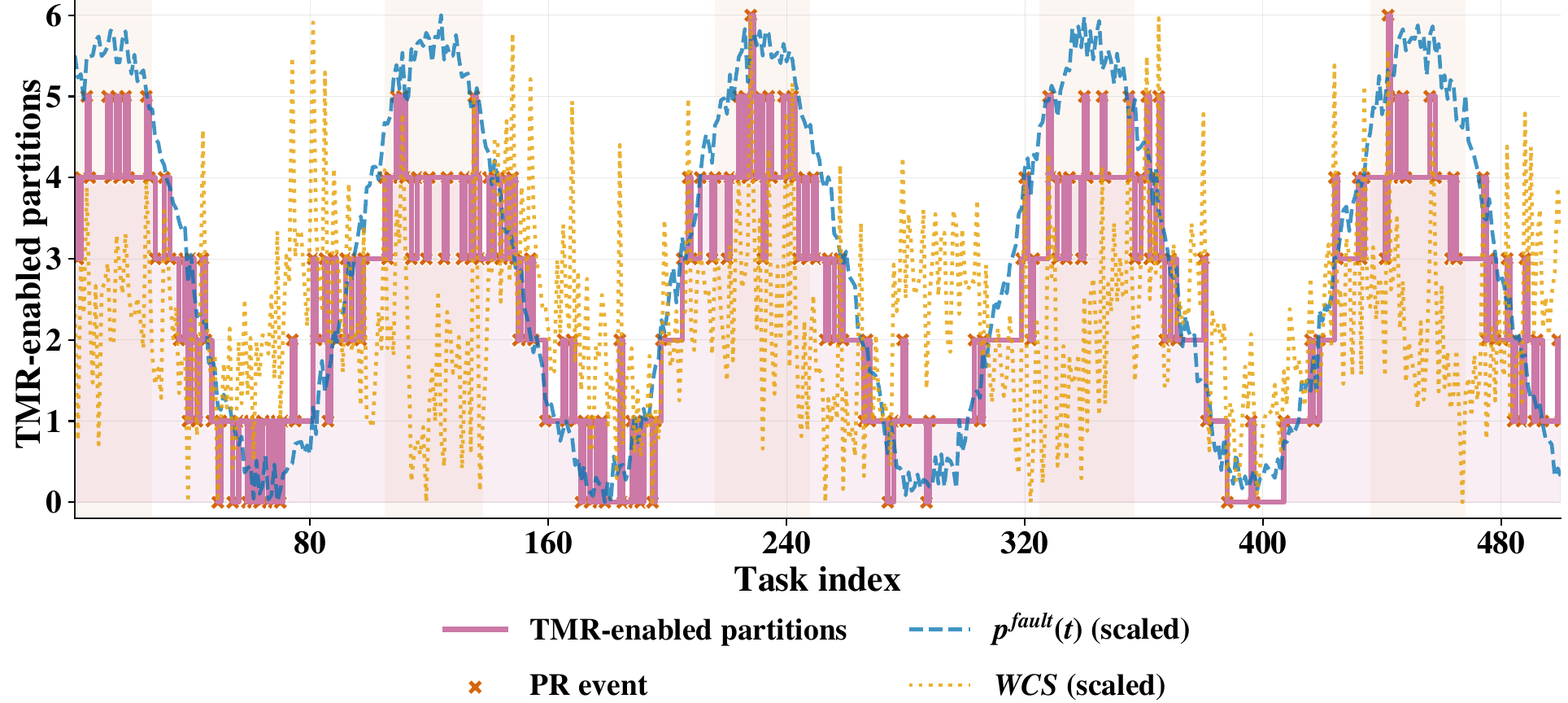}
\caption{Adaptive protection behavior over time. Example visualization: number of TMR-enabled partitions (0--6) per task, with PR events marked. The reactive baseline typically responds after fault detection, while ProWAFT adjusts proactively.}
\label{fig:adapt_timeline}
\end{figure}

Table~\ref{tab:adaptivity_analysis} quantifies ProWAFT's protection behavior across operational regimes. ProWAFT increases TMR usage as either fault risk or workload criticality rises, while maintaining stable sub-millisecond decision latency. Fig.~\ref{fig:adapt_timeline} provides a timeline view of the adaptive decisions over the trace.

\begin{table}[htbp]
\centering
\caption{ProWAFT Adaptation Behavior Across Operational Regimes}
\label{tab:adaptivity_analysis}
\begin{tabular}{@{}lcccc@{}}
\toprule
\textbf{Operational Regime} & \textbf{Avg $p^{fault}$} & \textbf{Avg $WCS$} & \textbf{TMR Usage (\%)} & \textbf{Avg Decision Latency (ms)} \\
\midrule
Low Risk & 0.002 & 0.25 & 12.3 & 0.42 \\
Moderate Risk & 0.005 & 0.45 & 34.7 & 0.47 \\
High Risk & 0.009 & 0.70 & 78.9 & 0.51 \\
Critical Workloads Only & 0.003 & 0.80 & 95.2 & 0.38 \\
\addlinespace
\textbf{Reactive-Reconfig Baseline} & -- & -- & 45.6 & 15.3 (post-fault) \\
\bottomrule
\end{tabular}
\end{table}

\subsubsection{Overhead (RQ3)}
Table~\ref{tab:overhead_breakdown} breaks down decision and reconfiguration overhead. The total decision overhead is 0.50\,ms (0.13\,mJ), dominated by candidate evaluation and WCS computation. PR dominates proactive overhead. In the single-partition case, proactive overhead (decision+PR) is 4.70\,ms, versus 15.30\,ms for reactive recovery (3.3$\times$ lower). Even for multi-partition updates (9.00\,ms), proactive overhead remains lower than reactive recovery.\cite{zhang2026memmark, chen2025r2i, chen2026mvibench, you2026drdgrl, zhao2026stride,huang2026gui}

\begin{table}[htbp]
\centering
\caption{ProWAFT Decision and Reconfiguration Overhead Breakdown}
\label{tab:overhead_breakdown}
\begin{tabular}{@{}lcc@{}}
\toprule
\textbf{Component} & \textbf{Time Cost} & \textbf{Energy Cost} \\
\midrule
Workload Feature Extraction & 0.08 ms & 0.02 mJ \\
WCS Computation (Eq.~\ref{eq:wcs}) & 0.12 ms & 0.03 mJ \\
Candidate Configuration Evaluation & 0.25 ms & 0.07 mJ \\
Policy Decision & 0.05 ms & 0.01 mJ \\
\addlinespace
\textbf{Total Decision Overhead} & \textbf{0.50 ms} & \textbf{0.13 mJ} \\
\addlinespace
Single Partition Reconfiguration & 4.20 ms & 1.10 mJ \\
Multi-Partition Reconfiguration (avg) & 8.50 ms & 2.25 mJ \\
\addlinespace
\textbf{Total Proactive Overhead (Decision+PR)} & \textbf{4.70--9.00 ms} & \textbf{1.23--2.38 mJ} \\
\textbf{Reactive Overhead (RR Baseline)} & \textbf{15.30 ms} & \textbf{4.05 mJ} \\
\bottomrule
\end{tabular}
\end{table}

\subsubsection{Component contribution (RQ4)}
Table~\ref{tab:ablation_results} reports ablations. Removing WCS increases $C_{total}$ and reduces success rate, showing that uniform criticality is insufficient. Removing the propagation model (FPF/RRS) reduces success rate (94.3\%), despite similar throughput. Replacing the policy with a greedy strategy leads to the largest $C_{total}$ increase (0.66).

\begin{table}[htbp]
\centering
\caption{Ablation Study: Impact of Individual Components on System Performance}
\label{tab:ablation_results}
\begin{tabular}{@{}lcccc@{}}
\toprule
\textbf{Method Variant} & \textbf{Norm. Throughput} & \textbf{Energy (J)} & \textbf{Success Rate (\%)} & $\mathbf{C_{total}}$ \\
\midrule
\textbf{Full ProWAFT} & \textbf{0.89} & \textbf{210.7} & \textbf{98.8} & \textbf{0.54} \\
\addlinespace
w/o WCS (uniform criticality) & 0.85 & 228.3 & 95.1 & 0.63 \\
w/o FPF (no propagation model) & 0.88 & 215.4 & 94.3 & 0.61 \\
w/o policy (greedy) & 0.82 & 225.6 & 97.2 & 0.66 \\
\addlinespace
\textbf{Degradation vs. Full} & -7.9\% to +8.3\% & +2.2\% to +8.3\% & -4.6 to -1.6 pp & +13.0\% to +22.2\% \\
\bottomrule
\end{tabular}
\end{table}

\subsubsection{Sensitivity analysis}
Table~\ref{tab:sensitivity_analysis} studies sensitivity to objective weights and estimation errors. Shifting weights moves the operating point as expected, while moderate perturbations in WCS threshold and fault probability estimates cause limited degradation, indicating stable behavior under reasonable modeling errors.

\begin{table}[htbp]
\centering
\caption{Sensitivity Analysis of ProWAFT to Key Parameters}
\label{tab:sensitivity_analysis}
\begin{tabular}{@{}lccc@{}}
\toprule
\textbf{Parameter Variation} & $\mathbf{C_{total}}$ & \textbf{Success Rate (\%)} & \textbf{Reconfig. Events} \\
\midrule
Baseline ($\eta_T{=}0.4,\eta_E{=}0.3,\eta_R{=}0.3$) & 0.54 & 98.8 & 127 \\
\addlinespace
Performance-focused ($\eta_T{=}0.6,\eta_E{=}0.3,\eta_R{=}0.1$) & 0.48 & 96.2 & 89 \\
Energy-focused ($\eta_T{=}0.3,\eta_E{=}0.6,\eta_R{=}0.1$) & 0.52 & 96.8 & 103 \\
Reliability-focused ($\eta_T{=}0.2,\eta_E{=}0.2,\eta_R{=}0.6$) & 0.58 & 99.5 & 156 \\
\addlinespace
WCS Threshold +20\% & 0.57 & 97.1 & 112 \\
WCS Threshold -20\% & 0.55 & 99.2 & 143 \\
\addlinespace
Fault Probability Estimate +25\% & 0.56 & 99.1 & 138 \\
Fault Probability Estimate -25\% & 0.58 & 97.9 & 115 \\
\bottomrule
\end{tabular}
\end{table}

\subsection{Summary of findings}
Table~\ref{tab:summary_findings} summarizes the answers to the four research questions.

\begin{table}[htbp]
\centering
\caption{Summary of Experimental Findings for Each Research Question}
\label{tab:summary_findings}
\begin{tabularx}{\linewidth}{@{}p{0.18\linewidth}X p{0.30\linewidth}@{}}
\toprule
\textbf{RQ} & \textbf{Key Finding} & \textbf{Supporting Evidence} \\
\midrule
Q1: Effectiveness &
ProWAFT achieves the best overall trade-off, reducing $C_{total}$ by 33.3\% vs.\ Static-TMR and 30.8\% vs.\ RR. &
Table~\ref{tab:overall_results}. \\
\addlinespace
Q2: Adaptivity &
ProWAFT adjusts TMR usage from 12.3\% to 95.2\% based on $WCS$ and $p^{fault}$. &
Table~\ref{tab:adaptivity_analysis}, Fig.~\ref{fig:adapt_timeline}. \\
\addlinespace
Q3: Overhead &
Online decision overhead is 0.50\,ms; proactive (decision+PR) overhead is lower than reactive recovery. &
Table~\ref{tab:overhead_breakdown}. \\
\addlinespace
Q4: Contribution &
WCS, propagation modeling, and policy are all necessary; removing any increases $C_{total}$ by 13--22\%. &
Table~\ref{tab:ablation_results}. \\
\bottomrule
\end{tabularx}
\end{table}

\section{Limitations}
Our experiments employ software-based fault injection, which enables controlled evaluation but cannot capture all physical mechanisms. Future work will include radiation testing and thermal-stress testing, and will reduce reliance on offline characterization by enabling in-field self-calibration.

\section{Conclusion}
This paper presented \textbf{ProWAFT}, a proactive and workload-aware fault-tolerance framework for FPGA-based CNN accelerators. ProWAFT leverages partial reconfiguration to selectively enable TMR based on workload criticality and time-varying fault risk, while explicitly accounting for reconfiguration overhead in the decision objective. Experiments on a Zynq UltraScale+ ZCU104 platform with six reconfigurable partitions and a 500-task CNN-layer trace show that ProWAFT achieves a better overall performance--energy--reliability trade-off than static redundancy and reactive recovery, with sub-millisecond online decision overhead. Future work will validate the approach under physical fault campaigns (e.g., radiation/thermal stress) and extend the framework to larger accelerator libraries and longer-running deployments.

\bibliographystyle{plainnat}
\bibliography{references}
\end{document}